# VyAnG-Net: A Novel Multi-Modal Sarcasm Recognition Model by Uncovering Visual, Acoustic and Glossary Features

Ananya Pandey, Dinesh Kumar Vishwakarma[*]

Biometric Research Laboratory, Department of Information Technology, Delhi Technological University, Bawana Road, Delhi-110042, India

ananyaphdit08@gmail.com, dvishwakarma@gmail.com[*]

**Abstract**

Various linguistic and non-linguistic clues, such as excessive emphasis on a word, a shift in the tone of voice, or an awkward expression, frequently convey sarcasm. The computer vision problem of sarcasm recognition in conversation aims to identify hidden sarcastic, criticizing, and metaphorical information embedded in everyday dialogue. Prior, sarcasm recognition has focused mainly on text. Still, it is critical to consider all textual information, audio stream, facial expression, and body position for reliable sarcasm identification. Hence, we propose a novel approach that combines a lightweight depth attention module with a self-regulated ConvNet to concentrate on the most crucial features of visual data and an attentional tokenizer-based strategy to extract the most critical context-specific information from the textual data. The following is a list of the key contributions that our experimentation has made in response to performing the task of Multi-modal Sarcasm Recognition: an attentional tokenizer branch to get beneficial features from the glossary content provided by the subtitles; a visual branch for acquiring the most prominent features from the video frames; an utterance-level feature extraction from acoustic content and a multi-headed attention based feature fusion branch to blend features obtained from multiple modalities. Extensive testing on one of the benchmark video datasets, MUSTaRD, yielded an accuracy of **79.86%** for speaker dependent and **76.94%** for speaker independent configuration demonstrating that our approach is superior to the existing methods. We have also conducted a cross-dataset analysis to test the adaptability of VyAnG-Net with unseen samples of another dataset MUStARD++.

*Keywords- Sarcasm, Multi-modal Sarcasm Recognition (MSR), Self-Regulated ConvNet, Glossary, Acoustic, Visual*

1. **Introduction**

The proposed model has been named "VyAnG," which draws inspiration from the use of sarcasm in the Hindi language. The acronyms V, A, and G correspond to visual, acoustic, and glossary (textual) content, respectively. The Multi-modal sarcasm recognition (MSR) task aims to determine whether a given video utterance should be labelled as sarcastic or non-sarcastic based on its content. Multi-modal learning has emerged as a significant area of study in recent years [1], [2] due to the proliferation of video clips and other user-generated multi-modal content on social networking sites. In contrast to conventional single-modal learning on individual modalities (such as auditory, visual, or textual), multi-modal learning seeks to combine multiple data streams into a single unit. MSR is a subset of multi-modal sentiment recognition in which the speaker deliberately uses unconventional body language, word choice, or vocal inflexion to emphasize incongruity across modalities. Unfortunately, it is notoriously tricky for opinion-mining algorithms to comprehend sarcastic utterances accurately. Consider the phrase "*Unconditionally; I love it whenever my train is delayed"* as an example. It is quite easy to be misled by the positive word *"love"* while failing to understand the underlying emotion from *"my train is late"*. Therefore, developing an accurate system for spotting sarcasm is quite vital when it comes to opinion mining.

This form of implicit human attitude is essential for preventing discussion barriers, producing positive effects on mental health, and fostering trust. On the other hand, existing deep learning architectures might sometimes struggle to understand such complex and multi-modal emotions.

### 1.1  Motivation

Text has always been the most prevalent medium for delivering sarcasm. Sarcasm in multi-modal data, on the other hand, typically requires explicit inter-modal clues to expose the speaker's true intent. It may, for example, be indicated by a mix of linguistic and non-linguistic clues, such as excessive emphasis on a phrase, a drawn-out syllable, a shift in tone of voice, or an awkward expression. Consider the cheering statement in **Figure 1**: *"if you're compiling a mix CD for a double suicide. Oh, I hope that scratching post is for you."* becomes sarcastic when spoken with an awkward face and a saucy tone, and in general, has a negative meaning. Naturally, humans can process this massive amount of simultaneous data. However, developing an approach that can possibly accomplish the same task requires a suitable representation of all of these different sources of information. It thus results in a significant increase in research interest.

*Significance of the dataset used in this study:* In today's era, individuals are expressing their viewpoints on social media via the use of sarcastic modes of communication. Furthermore, there is a notable trend of people actively participating in the dissemination of caustic reviews directed towards a wide range of political decisions and governmental policies. Hence, apart from understanding of multiple emotions, the ability to detect sarcasm is crucial for individuals to successfully deal with and engage in modern societies that are flooded with irony. Considering the significance of sarcasm recognition, we decided to choose a multimodal dataset for sarcasm recognition to assess the effectiveness our proposed approach.

### 1.2  Challenges

Sarcastic dialogue utterances are challenging to gather, despite their frequent use in movies and television series, due to the time and effort needed to recognize and annotate unprocessed videos with sarcastic class labels manually. MUStARD, the sole open-source accessible dataset, has 690 video clips labelled with either non-sarcastic or sarcastic sentiments. The main challenge in the MSR field is acquiring the most prominent features from all the modalities. Thus, to obtain robust intra-modal dependencies, we offer a novel framework integrating a dedicated lightweight depth attention module in the visual branch to extract the most prevalent features from the video frames. In contrast, the textual branch uses the attention-based tokenization method to acquire the most relevant features from the glossary content provided by the subtitles and to acquire inter-modal dependencies; in our study, the proposed approach employs multi-headed attention fusion to integrate asynchronously received features from each of the separate modalities.

## 1.3 Major Contributions

The following are the three aspects of our contribution:

- We proposed VyAnG-Net, a novel multi-modal sarcasm recognition framework, by uncovering visual, acoustic and glossary (textual) features. This framework includes the glossary branch that uses the attention-based tokenization approach to acquire the most significant contextual features from the textual content provided by the subtitles of the video utterances, a visual unit with a dedicated lightweight depth attention module to acquire the most prominent features from the video frames, an utterance-level feature extraction from acoustic content and lastly multi-headed attention based feature fusion has been employed to blend features acquired from each of the separate modalities.
- We have tested our method on one of the standard video datasets, MUStARD, and found that it surpasses the cutting-edge techniques by a wide margin.
- Finally, a series of ablation experiments were conducted to ensure the robustness of our proposed approach that we suggested, and our results seemed quite promising.
- In the recent years, remarkable advancements have been made in sarcasm identification frameworks on the MUStARD dataset, but there are concerns about their generalizability. Consequently, rather than limiting ourselves to assessing VyAnG-Net on a single dataset we undertake a cross-dataset study as part of a generalization research to test the resilience of VyAnG-Net. Our proposed approach, VyAnG-Net, was trained using the MUStARD dataset for this experimental investigation, and its performance was evaluated using an unseen MUStARD++ dataset.

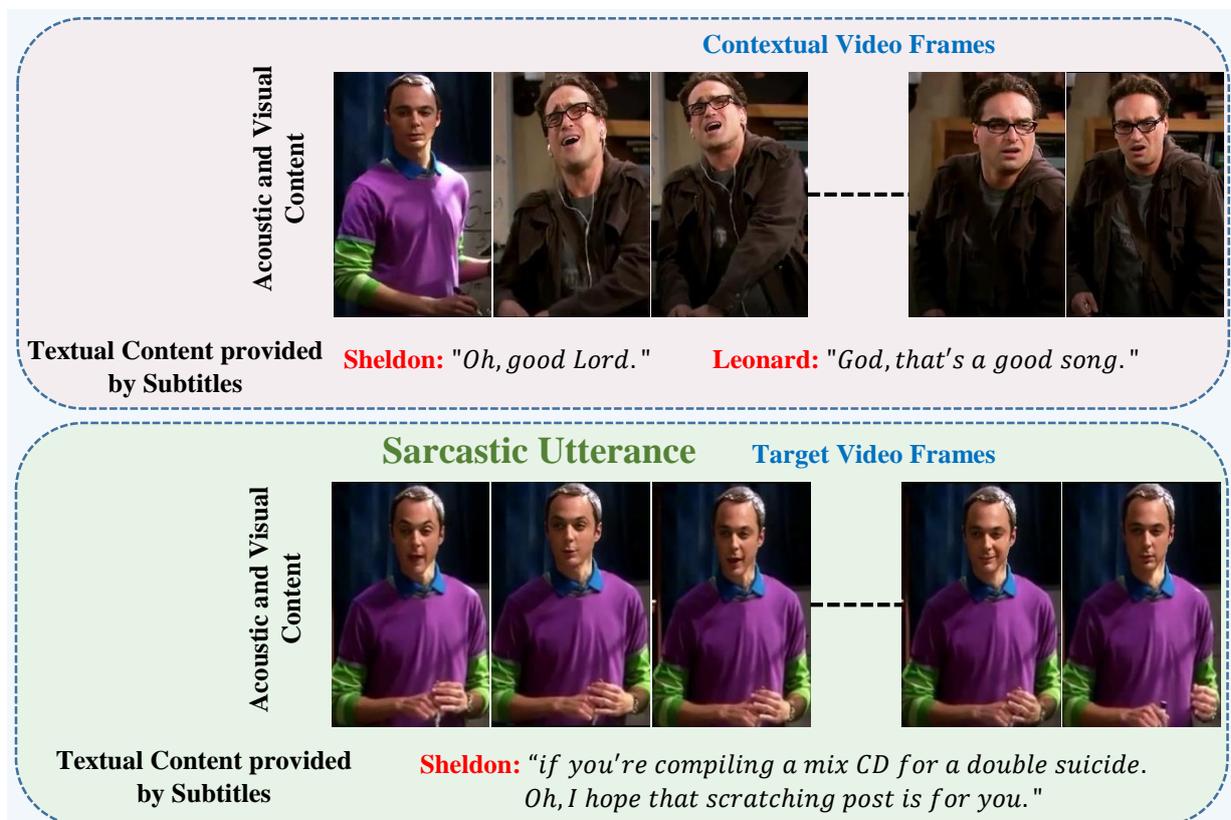

**Figure 1** A perfect illustration of a sarcastic statement from the dataset, accompanied by its context and a transcript of the utterance

Previous studies have employed traditional machine learning methods to investigate multimodal sarcasm identification in their research. We have discovered a revolutionary blend of advanced deep-learning frameworks that is highly suitable for this task. The incorporation is unique and yields the most optimal outcomes compared to the baselines.

## 2    Related Work

This section deals with both single-modal and multi-modal sarcasm recognition cutting-edge approaches from the previous research studies in detail.

### 2.1    Unimodal Sarcasm Recognition

The current subsection covers literature on sarcasm recognition that focuses on a single modality in their research.

#### 2.1.1    Sarcasm Recognition through Text

Most prior research for identifying sarcasm in text-based content has focused on either lexicon or rule-based approaches. Twitter is used as a leading data source for resource collection in this area, with human annotations [3], [4] and remote supervision through hashtags serving as the key annotating methods [5], [6]. According to previous studies, similarities between a speaker and the audience may also be gleaned from their context [7]. Apart from this, many other different aspects of context have also been addressed, such as the encoding of expressed emotion and the individual's personality traits [8]; the background of the speaker and behavioural patterns on various online platforms [9]; aspects of style and discourse [10]; features of the user community [11]; and the retention of user-specific representations [11], [12].

Traditionally, classical machine learning approaches have been utilized to recognise sarcasm. Generally, algorithms like Naive Bayes, SVM, logistic regression, etc., were used for various text classification tasks, as indicated in [13], [14]. These approaches deliver the most accurate and effective results when dealing with classification challenges involving smaller datasets. Furthermore, in [15], a comparison of conventional and transfer learning strategies was analysed. In most scenarios, ensemble-based machine learning approaches also provide the best performance for text-based classification in detecting sarcasm, as presented in [16].

Machine learning methods will be insufficient to deal with the massive amounts of data generated daily on social media platforms. Hence, in addition, to typical machine learning algorithms, researchers are increasingly more enthusiastically working on sarcasm identification using different deep learning approaches to manage vast amounts of information. For example, to recognize sarcasm on the web, [17] utilizes numerous deep learning architectures, including GRU, LSTM, and ConvNets, to form an ensemble-based model. In the aviation industry, RNNs with GRU and SVM were utilized [18] to enhance the recognition and analysis of sarcastic sentiments. In addition to using English as the sole language for the research, other languages, such as Hindi and Arabic, have also been utilized to identify sarcasm in tweets. [19] suggested *"TANA"*, a neural architecture to recognise sarcasm in Hindi tweets. The system is trained using word and emoji embedding and uses a combination of LSTM and SVM to identify sarcasm.

Transformer-based models are also receiving a lot of attention because of their self-attention mechanism [20], which aids in focusing on the most salient characteristics while ignoring the rest for various computer vision and natural language processing applications. [21] proposed a hybrid deep neural architecture with an integrated attention module for identifying sarcasm in news headlines. A combination of graph-convolution neural network and BERT has been presented [22] for recognizing irony in the text.

### 2.1.2 Sarcasm Recognition through Audio

The acquisition of prosodic signals in the form of auditory patterns that are associated with sarcastic behaviour has been the primary focus for recognizing sarcasm in voice. For addressing the issue of sarcasm detection using acoustic content was proposed by [23], who focused on the vocal tonalities of ironical speech. The researchers speculated that slower speech rates and greater frequency could be the best indicators of sarcasm. Researchers [24] looked at prosodic and spectral aspects of sound in and out of context to identify sarcasm. Stress & intonation are two examples of prosodic characteristics that are widely regarded as reliable predictors of irony [25]. The above-discussed research contributions are among the few discussions devoted to sarcasm detection based on audio.

## 2.2 Multi-Modal Sarcasm Recognition

This subsection examines the research studies on sarcasm recognition that focuses on multiple modalities.

### 2.2.1 Sarcasm Recognition through Image-Text Pairs

Affective computing is currently attracting a lot of interest from academics, particularly with regard to the usage of multi-modal sources of information. People are still heavily dependent on text-based content, but videos, acoustic, images, and emoticons are becoming more popular nowadays. Consequently, we frequently encounter posts on online community forums that include text and a caption. Hence, sarcasm recognition using image-text pairings has been the subject of extensive study. [26] was the first to work on multi-modal sarcasm identification using image-text pairings scraped from Instagram posts. Bi-GRU was used in this research study to extract features from captions, while VGG-16 was used to extract features from the visual modality. In addition, the text included in the picture is retrieved using OCR, and the resulting transcript is given to the Bi-GRU for feature extraction. Finally, all of the cross-modality features are concatenated and sent to a classification layer to provide the final prediction score. [27] have also examined the interaction of textual and visual information in sarcastic multi-modal blogs for three prominent social media sites, namely, Tumblr, Instagram & Twitter, and present a classification of the relevance of photos in sarcastic posts. Researchers have recently started to employ "attention modules" to help them focus on what's important and ignore the rest in order to acquire more accurate and superior results. For example, [28] suggested a deep learning framework based on attention mechanisms to accomplish multiple subtasks such as sarcasm, sentiment, and humour recognition for image-text pairs. [29]–[31] are other contributions in recent years based on image-text pairs.

### 2.2.2 Sarcasm Recognition through Videos

Despite the fact that multi-modal data sources provide extra clues in identifying sarcasm, this has not been done extensively; one of the primary reasons behind this is the lack of multi-modal datasets. In the modern era, scholars [32] [33] have commenced utilising diverse sources of information to detect sarcasm, including those that are multi-modal in nature. Indeed, it is a verifiable fact that modalities such as acoustic and visual frequently offer a greater abundance of clues pertaining to the context of an utterance when compared to text. The MUStARD dataset, which is the initial multi-modal video dataset for detecting sarcasm, was recently introduced by [32]. In this study, researchers have used an SVM algorithm to identify examples of sarcasm in the given dataset. Subsequently, the MUStARD dataset was manually annotated with sentiment and emotion class labels by [34], followed by sentiment and emotion recognition, along with sarcasm recognition. [35] introduces *"IWAN"*, an innovative approach for identifying sarcasm. This method involves the utilisation of a scoring mechanism that prioritises word-level incongruity details. The task of analysing emotions in sarcastic video utterances has evolved with the release of MUStARD++, an extended version of MUStARD that includes nine distinct emotions. This development was reported by [36] with nine emotions for the task of emotion analysis in video utterances.

In addition to deep learning methodologies, a novel approach utilising fuzzy logic has been proposed for the recognition of sarcasm in video utterances by [37]. The SEEmoji MUStARD dataset has been recently published by [38] as an extension of the benchmark dataset MUStARD. This dataset is intended for the analysis of sentiment, sarcasm, and emotion and utilises a multi-task framework that is capable of recognising emojis as well.

Although previous research endeavours have had significant efforts on the alignment of visual, acoustic and textual features, however, the acquisition of multi-modal sarcastic video samples poses a significant challenge, and currently existing approaches encounter difficulties in achieving satisfactory performance levels on the MUStARD dataset. This served as a source of inspiration for us to come up with a framework that can effectively incorporate key information from all three modalities in the context of multi-modal sarcasm recognition.

## 3  Proposed Approach

This section provides a comprehensive discussion of the proposed framework VyAnG-Net. The main objective is outlined in the first part of this section. Then, the model's framework is introduced, which consists of three modules: a glossary branch that uses the attention-based tokenization approach to acquire the most significant contextual features from the textual content provided by the subtitles of the video utterances, a visual branch with dedicated attention module to acquire the most prominent features from the video frames and lastly multi-headed attention based feature fusion to blend features acquired from each of the separate modalities. The term "multi-modal sarcasm recognition" is typically used in our study to denote the process of analysing sarcasm in video utterances for the sake of convenience.

### 3.1  Objective

The problem of recognising sarcasm in video utterances can be thoroughly summed up as follows:

Let "$\mathcal{V}$" denote the sample space comprising video utterances. A sample of the dataset consists of the textual content conveyed through video subtitles "$\mathbb{G}$", visual frames "$\mathbb{V}$", and accompanying acoustic content "$\mathbb{A}$". Each of the samples is assigned to a class label denoted as "$\mathbb{C}$". In more technical terms, each sample can be defined as a quartet consisting of a subtitle, visual frames, acoustic information, and a class label. The following expression can be formulated as:

$$\mathbb{I} = \{(\mathbb{G}^0, \mathbb{V}^0, \mathbb{A}^0, \mathbb{C}^0), (\mathbb{G}^1, \mathbb{V}^1, \mathbb{A}^1, \mathbb{C}^1), \ldots, (\mathbb{G}^i, \mathbb{V}^i, \mathbb{A}^i, \mathbb{C}^i), \ldots, (\mathbb{G}^{m-1}, \mathbb{V}^{m-1}, \mathbb{A}^{m-1}, \mathbb{C}^{m-1})\} \qquad (1)$$

where, $\mathbb{I}$ is the collection of sample quartets, $\mathbb{G}^i$ denotes subtitle information, $\mathbb{V}^i$ denotes visual frame information, $\mathbb{A}^i$ defines the acoustic content, $\mathbb{C}^i$ is the class label that corresponds to a specific utterance for the $i^{th}$ sample and the variable $m$ represents the cardinality of the sample space, which denotes the total number of samples in a given dataset.

VyAnG-Net aims to learn a mapping function $\mathbb{F}:(\mathbb{G},\mathbb{V},\mathbb{A}) \rightarrow \mathbb{C}$ from the multi-modal training examples $\{(\mathbb{G}^i, \mathbb{V}^i, \mathbb{A}^i) | 0 \leq i \leq m-1\}$. For a sarcasm recognition task, $\mathbb{C}^i \in \{sarcasm\ and\ not\ sarcasm\}$.

### 3.2 VyAnG-Net: A Novel Multi-Modal Sarcasm Recognition Model by Uncovering Visual, Acoustic and Glossary Features

The VyAnG-Net was proposed for the purpose of performing multi-modal sarcasm recognition to generate the relationship among visual, acoustic, and glossary (textual) information and to explore the compatibility between these three modalities. **Figure 2** illustrates the VyAnG-Net framework. The model comprises of three distinct components. Firstly, a textual branch that employs an attention-based tokenization approach to extract the most salient contextual features from the glossary content presented in the video utterances' subtitles. Secondly, a visual branch that incorporates a dedicated attention module to capture the most prominent features from the video frames. Lastly, a multi-headed attention-based feature fusion mechanism is utilised to integrate the features obtained from each of the individual modalities. **Table 1** presents the proposed framework in algorithmic format.

**Table 1** Pseudocode for the proposed VyAnG-Net

| |
|---|
| **3.3 Algorithm 1:** VyAnG-Net: A Novel Multi-Modal Sarcasm Recognition Model by Uncovering Visual, Acoustic and Glossary Features. |
| **Aim:** To learn a mapping function $\mathbb{F}:(\mathbb{G},\mathbb{V},\mathbb{A}) \rightarrow \mathbb{C}$ from the multi-modal training examples $\{(\mathbb{G}^i, \mathbb{V}^i, \mathbb{A}^i) | 0 \leq i \leq m-1\}$.<br>Input: Glossary (textual) set $\mathbb{G} = \{\mathbb{G}_1, \mathbb{G}_2, \ldots, \mathbb{G}_i\}$, visual set $\mathbb{V} = \{\mathbb{V}_1, \mathbb{V}_2, \ldots, \mathbb{V}_i\}$, and acoustic set $\mathbb{A} = \{\mathbb{A}_1, \mathbb{A}_2, \ldots, \mathbb{A}_i\}$.<br>**Output:** sarcasm recognition task, $\mathbb{C}^i \in \{sarcasm\ and\ not\ sarcasm\}$. |
| 1. Word-to-vector representation from the entire Glossary content set $\mathbb{R}$;<br>2. Extract features at the level of utterance and context from vector representation of the glossary content $\mathbb{G}_u \oplus \mathbb{S}_{gu}$, and $\mathbb{G}_c \oplus \mathbb{S}_{gc}$<br>3. Extract features at the level of utterance and context from the visual frame $\mathbb{V}_u \oplus \mathbb{S}_{vu}$, and $\mathbb{V}_{uc} \oplus \mathbb{S}_{vc}$;<br>4. Extract features at the level of utterance and context from acoustic content $\mathbb{A}_u \oplus \mathbb{S}_{au}$, and [ ]<br>5. for $\mathbb{E} \leftarrow 1$ to $\mathbb{Epochs}$ do<br>    $\mathbb{R}_u \leftarrow \mathbb{W}_{1:g} = \{\mathbb{W}_1, \mathbb{W}_2, \ldots, \mathbb{W}_g\}$ word to vector representation by **Eq. $\langle 1 \rangle$**;<br>    $\mathbb{R}_c \leftarrow \mathbb{W}_{1:g_c^i} = \{\mathbb{W}_1^i, \mathbb{W}_2^i, \ldots, \mathbb{W}_{g_c}^i\}$ word to vector representation by **Eq. $\langle 4 \rangle$**;<br>    $\mathbb{G}_u \oplus \mathbb{S}_{gu} \leftarrow (\mathbb{G}_u \leftarrow \mu(\mathbb{R}_u) \oplus \mathbb{S}_{gu})$ obtain utterance-level text-based features using **Eq. $\langle 2 \rangle$** and $\langle 3 \rangle$;<br>    $\mathbb{G}_c \oplus \mathbb{S}_{gc} \leftarrow (\mathbb{G}_c^i \oplus \mathbb{S}_{gc}^i)$ obtain context-level text-based features using **Eq. $\langle 2 \rangle$** and $\langle 3 \rangle$;<br>    $\mathbb{V}_u \oplus \mathbb{S}_{vu} \leftarrow \mathbb{VS}_u$ obtain utterance-level visual features using **Eq. $\langle 9 \rangle$** and $\langle 10 \rangle$;<br>    $\mathbb{V}_{uc} \oplus \mathbb{S}_{vc} \leftarrow \mathbb{VS}_c$ obtain context-level visual features using **Eq. $\langle 11 \rangle$** and $\langle 12 \rangle$;<br>    $\mathbb{A}_u \oplus \mathbb{S}_{au} \leftarrow$ **Librosa tool**($Concatenation(\mathbb{A}_u, \mathbb{S}_{au})$) obtain utterance-level acoustic features using **Eq. $\langle 13 \rangle$**;<br>    $\mathbb{G}_{cat} \leftarrow \mathbb{G}_u \oplus \mathbb{S}_{gu} \oplus \mathbb{G}_c \oplus \mathbb{S}_{gc}$ final text-based features obtained by concatenating utterance and context-based features;<br>    $\mathbb{V}_{cat} \leftarrow \mathbb{V}_u \oplus \mathbb{S}_{vu} \oplus \mathbb{V}_{uc} \oplus \mathbb{S}_{vc}$ final vision-based features obtained by concatenating utterance and context-based features;<br>    $\mathbb{A}_{cat} \leftarrow \mathbb{A}_u \oplus \mathbb{S}_{au} \oplus [\ ]$ final audio-based features obtained by concatenating utterance and context-based features;<br>    $\mathbb{L}_\mathbb{G} \leftarrow \mathbb{G}_{cat}$ obtain the most prominent textual features by applying multi-headed attention using **Eq. $\langle 15 \rangle$**;<br>    $\mathbb{L}_\mathbb{V} \leftarrow \mathbb{V}_{cat}$ obtain the most prominent visual features by applying multi-headed attention using **Eq. $\langle 15 \rangle$**;<br>    $\mathbb{L}_\mathbb{A} \leftarrow \mathbb{A}_{cat}$ obtain the most prominent acoustic features by applying multi-headed attention using **Eq. $\langle 15 \rangle$**;<br>    $\mathbb{GVA} \leftarrow \mathbb{L}_\mathbb{G} \oplus \mathbb{L}_\mathbb{V} \oplus \mathbb{L}_\mathbb{A}$ concatenate all the features obtained from multiple modalities to get multi-modal feature representation using **Eq. $\langle 16 \rangle$**; |

$\mathbb{C} \leftarrow \boldsymbol{Softmax}(\mathbb{GVA})$ pass the multi-modal features to the softmax layer to get the final prediction;

calculate loss and perform backpropagation;

6. *End*

'⊕' denotes concatenation & '[ ]'denotes the empty list

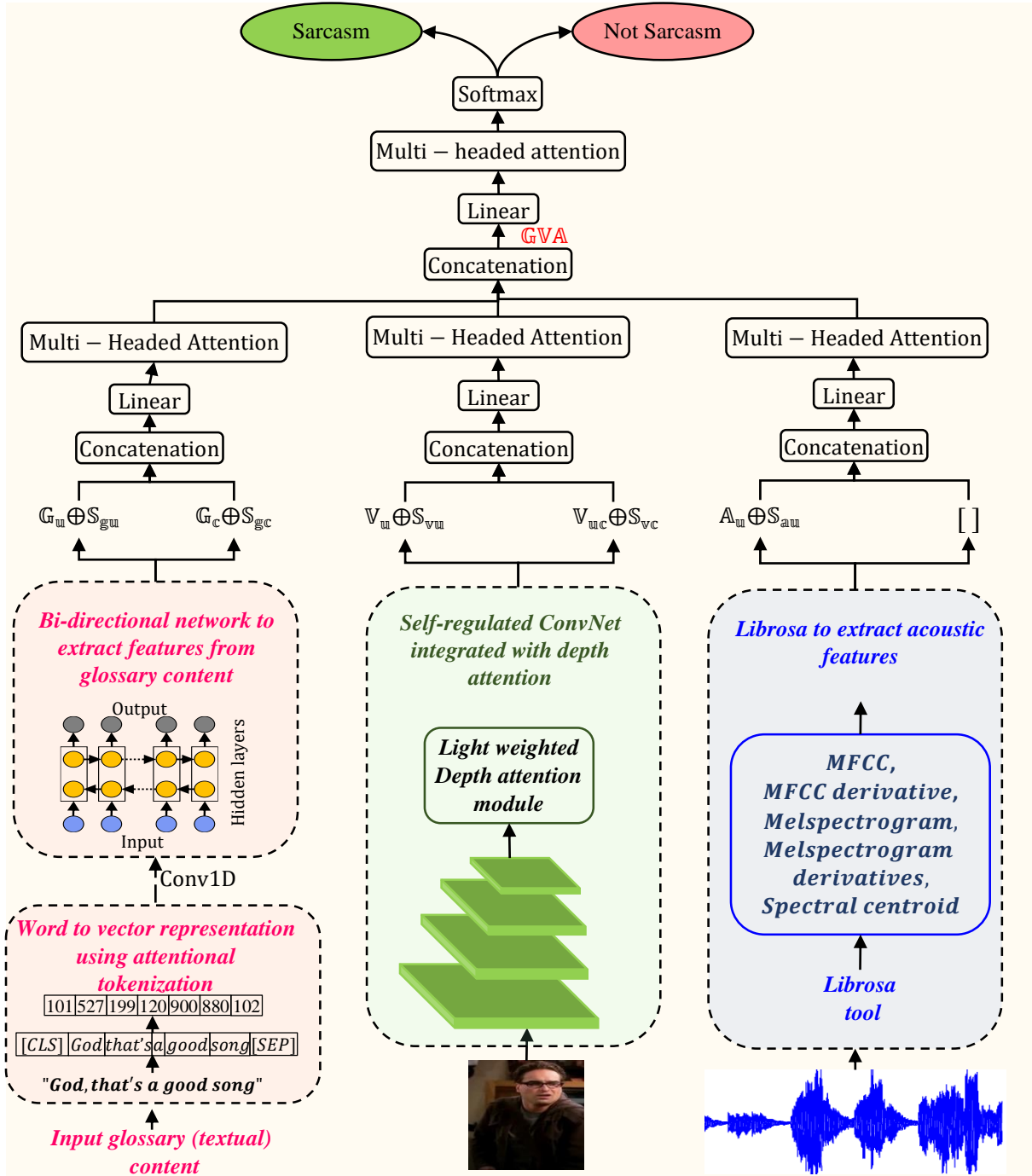

**Figure 2** Proposed VyAnG-Net Framework where, $\mathbb{G}, \mathbb{V}, \mathbb{A}$ corresponds to glossary (textual), visual, and acoustic content. The notations $\mathbb{G}_\mathbb{u} \oplus \mathbb{S}_{g\mathbb{u}}$, $\mathbb{V}_\mathbb{u} \oplus \mathbb{S}_{v\mathbb{u}}$, and $\mathbb{A}_\mathbb{u} \oplus \mathbb{S}_{a\mathbb{u}}$ refers to the utterance level features for all the three

modalities. Additionally, $\mathbb{G}_\mathbb{C} \oplus \mathbb{S}_{g\mathbb{C}}$, $\mathbb{V}_{u\mathbb{C}} \oplus \mathbb{S}_{v\mathbb{C}}$ represents the context level feature for glossary and visual content. Lastly, [ ] denotes the empty list.

### 3.3.1 Input Features

The dataset comprises of individual samples that include an utterance, its corresponding context, and associated labels. The utterance's context encompasses a series of prior utterances, typically $\mathbb{N}$ in number, that lead up to the given utterance within the dialogue. Each utterance is linked to its respective context and speaker, with the speaker of the utterance and the speaker of the context being distinct entities. Our study provides an extensive explanation of the utterance and its contextual factors across all modalities in the following subsections.

### 3.3.2 Textual Feature Extraction using Glossary Content

Assuming a given utterance consisting of $\mathbb{g}$ words, denoted as $\mathbb{W}_{1:\mathbb{g}} = \{\mathbb{W}_1, \mathbb{W}_2, \ldots, \mathbb{W}_\mathbb{g}\}$, where each word $\mathbb{W}_i$ belongs to the set of real numbers $\Re^{300}$. Each term is denoted as $\mathbb{W}_i$, corresponds to a vector that is generated through the utilization of [39] attention-based tokenization ($\tau$) represented in **Eq. ⟨1⟩**. The acquisition of the contextual relationship among words is accomplished by means of employing a [40] model denoted as $\mu$ using **Eq. ⟨2⟩**. Subsequently, utterance level features are obtained through the utilization of the final word embedding, represented as $\mathbb{G}_\mathbb{u}$.

$$\mathbb{R}_\mathbb{u} = \tau(\{\mathbb{W}_1, \mathbb{W}_2, \ldots, \mathbb{W}_\mathbb{g}\}) \qquad \langle 1 \rangle$$

$$\mathbb{G}_\mathbb{u} = \mu(\mathbb{R}_\mathbb{u}) \qquad \langle 2 \rangle$$

In cases where speaker information $\mathbb{S}_{g\mathbb{u}}$ is accessible, it is possible to combine it using **Eq. ⟨3⟩** with $\mathbb{G}_\mathbb{u}$ to form a speaker-aware textual utterance, which is represented $\mathbb{G}_\mathbb{u} \oplus \mathbb{S}_{g\mathbb{u}}$.

$$\mathbb{G}_\mathbb{u} \oplus \mathbb{S}_{g\mathbb{u}} = Concatenation(\mathbb{G}_\mathbb{u}, \mathbb{S}_{g\mathbb{u}}) \qquad \langle 3 \rangle$$

Assuming there is a set of utterances in the given context, each comprising $\mathbb{g}_\mathbb{c}$ words, the utterance-level representations for such a set of contextual videos are obtained by subjecting the words of each utterance to [40], using **Eq. ⟨4⟩** following which the embedding of the last word of the glossary provided by the subtitle is utilized. The $\mathbb{i}^{th}$ utterance in the context is denoted by $\mathbb{G}_\mathbb{C}^\mathbb{i}$.

$$\mathbb{G}_\mathbb{C}^\mathbb{i} = \mu\left(\tau\left(\{\mathbb{W}_{1:\mathbb{g}_\mathbb{C}^\mathbb{i}} = \{\mathbb{W}_1^\mathbb{i}, \mathbb{W}_2^\mathbb{i}, \ldots, \mathbb{W}_{\mathbb{g}_\mathbb{C}^\mathbb{i}}^\mathbb{i}\}\}\right)\right) \qquad \langle 4 \rangle$$

When speaker information $\mathbb{S}_{g\mathbb{c}}^\mathbb{i}$ is available, it is appended to every contextual utterance $\mathbb{G}_\mathbb{C}^\mathbb{i}$ too. In the end, the features at the context level are also obtained by concatenating all textual utterances that are influenced by the speaker, as represented in **Eq. ⟨5⟩**.

$$\mathbb{G}_\mathbb{C} \oplus \mathbb{S}_{g\mathbb{C}} = Concatenation\left((\mathbb{G}_\mathbb{C}^1 \oplus \mathbb{S}_\mathbb{C}^1), (\mathbb{G}_\mathbb{C}^2 \oplus \mathbb{S}_\mathbb{C}^2), \ldots, (\mathbb{G}_\mathbb{C}^\mathbb{i} \oplus \mathbb{S}_{g\mathbb{C}}^\mathbb{i})\right) \qquad \langle 5 \rangle$$

### 3.3.3 Visual Feature Extraction from the Video Frames of the Utterances

To obtain visual features from the video frames [41] is integrated with the depth attention module [42], discussed in the following section.

#### 3.3.3.1 Lightweight Attention Module

The Convolutional Neural Networks (ConvNets) have demonstrated remarkable representational abilities, leading to significant enhancements in their efficacy for visual tasks.

In addition, we explore another aspect of architectural design that has become increasingly prevalent in modern times, namely, attention. Through the utilization of attention mechanisms, which involve prioritising significant attributes while inhibiting irrelevant ones, it is anticipated that the efficacy of representation will be enhanced. Considering this information, a framework known as the "light weighted attention framework" [42], illustrated in **Figure 3**, has been developed and integrated into [41] to concentrate on the most salient characteristics from the visual frames while disregarding the others. In order to accomplish this task, we have implemented four different modules, namely, feature grouping, depth attention, spatial attention and aggregation, which constitute [42].

The word "spatial" refers to the encompassing spatial domain of each feature map. By including the spatial attention module to enhance the feature maps, the superior input is then sent to the subsequent levels of convolution, hence increasing the efficacy of the model. On the other hand, the phrase "depth" denotes the total number of channels, which are simply a set of feature maps arranged in a tensor. Each and every multidimensional layer inside this tensor represents a feature map with a depth of $\mathbb{H} \times \mathbb{W}$. The depth attention mechanism provides a numerical value associated with every channel, therefore prioritising those channels that have the most impact on the learning process. This prioritisation leads to the optimisation of the most important features, ultimately enhancing the overall performance of the model.

The property of feature grouping is characterised by a hierarchical structure consisting of two levels. Suppose that the attention module's input tensor is $\mathbb{X} \in \mathbb{R}^{\mathbb{D} \times \mathbb{H} \times \mathbb{W}}$, where $\mathbb{D}, \mathbb{H}\ and\ \mathbb{W}$ denote the depth, height and width of the feature map, respectively. Initially, $\mathbb{X}$ is partitioned into $\mathbb{P}$ distinct groups, resulting in $\mathbb{X}' \in \mathbb{R}^{\frac{\mathbb{D}}{\mathbb{P}} \times \mathbb{H} \times \mathbb{W}}$ for each group across the depth of the feature maps. The obtained feature groups are then transmitted to the attention components, where they are subsequently segregated into two distinct groups based on the depth dimension. One group is allocated to the spatial attention branch, while the other is assigned to the depth attention branch. And these sub-feature groups that are transmitted across both the spatial or depth attention branches can be represented as $\mathbb{X}'' \in \mathbb{R}^{\frac{\mathbb{D}}{2\mathbb{P}} \times \mathbb{H} \times \mathbb{W}}$.

The depth attention branch involves reducing the feature maps obtained from the feature grouping phase to $\mathbb{X}'' \in \mathbb{R}^{\frac{\mathbb{D}}{2\mathbb{P}} \times \mathbf{1} \times \mathbf{1}}$. This is achieved through the use of a global average pooling operation and gating mechanism, which enables more accurate and versatile decisions. The resulting output is then subjected to a sigmoid activation function which is represented as follows:

$$X_{\Bbbk 1}^{\hat{}} = \sigma\big(\mathbb{F}_{\mathbb{c}}(\mathbb{t})\big) \cdot \mathbb{X}'' = \sigma(\mathbb{V}_{1\mathbb{t}} \oplus \mathbb{b}_1) \cdot \mathbb{X}'' \tag{6}$$

The Group Norm technique is employed to reduce the input $\mathbb{X}'$ in spatial attention, resulting in spatial features. The function $\mathbb{F}(.)$ is subsequently employed to improve the depiction of the diminished tensor. This concept can be expressed through a simple mathematical formula:

$$X_{\Bbbk 2}^{\hat{}} = \sigma\big(\mathbb{V}_2 \cdot GroupNorm(\mathbb{X}'')\oplus\mathbb{b}_2\big) \cdot \mathbb{X}'' \tag{7}$$

The concatenation of the outputs obtained from the Spatial Attention and depth attention is performed initially. Then a depth shuffle technique is implemented, similar to the approach used in ShuffleNet, to facilitate interaction among groups along the depth. Consequently, the resulting output possesses identical dimensions to those of the input tensor that underwent processing in the shuffle attention layer.

$$X_{\Bbbk}^{\hat{}} = \big[X_{\Bbbk 1}^{\hat{}} \oplus X_{\Bbbk 2}^{\hat{}}\big] \in \mathbb{R}^{\frac{\mathbb{D}}{\mathbb{P}} \times \mathbb{H} \times \mathbb{W}} \tag{8}$$

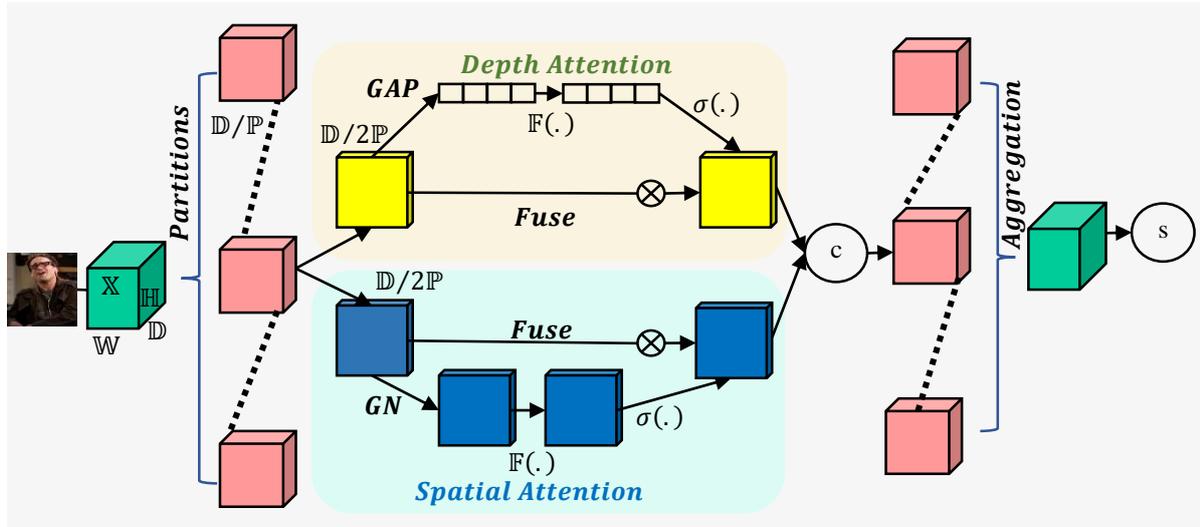

**Figure 3** Light weighted depth attention module where GAP represents global average pooling operation, GN represents group normalization, "C represents concatenation and "S" represents depth shuffle operation

#### 3.3.3.2 Utterance and Context-Level Feature Extraction from Video Frames

In this, utterance-level features are initially extracted, followed by the extraction of context-level features. These two sets of features are then subsequently concatenated to yield the final feature representation. Suppose there is a set of $\mathbb{N}_\mathbb{u}$ visual frames at utterance level denoted as $\mathbb{V}_{1:\mathbb{N}_\mathbb{u}} = \{\mathbb{V}_1, \mathbb{V}_2, \ldots\ldots, \mathbb{V}_{\mathbb{N}_\mathbb{u}}\}$. Each visual frame is sent to a self-regulatory ConvNet model [41] depicted in **Figure 4** that makes use of a light-weighted depth attention module [42] to extract the most prominent features, which has already been explained in detail above. To obtain information pertaining to the level of utterance, the mean value is computed for all frames $\mathbb{V}_\mathbb{u}$. In cases where speaker information is present, the utterance $\mathbb{V}_\mathbb{u}$ is concatenated with the corresponding speaker information $\mathbb{S}_{\mathbb{v}\mathbb{u}}$ given by **Eq. ⟨9⟩** and **Eq. ⟨10⟩**. The notation $\mathbb{V}_\mathbb{u} \oplus \mathbb{S}_{\mathbb{v}\mathbb{u}}$ used is where $\mathbb{V}_\mathbb{u}$ belongs to the set of real numbers $\Re$ and has a cardinality of 2048.

$$\mathbb{V}\mathbb{S}_\mathbb{u} = \big((\mathbb{V}_{\mathbb{u}1} \oplus \mathbb{S}_{\mathbb{v}\mathbb{u}1}), (\mathbb{V}_{\mathbb{u}2} \oplus \mathbb{S}_{\mathbb{v}\mathbb{u}2}), \ldots\ldots, (\mathbb{V}_{\mathbb{u}\mathbb{N}} \oplus \mathbb{S}_{\mathbb{v}\mathbb{u}\mathbb{N}})\big) \quad \langle 9 \rangle$$

$$\mathbb{V}_\mathbb{u} \oplus \mathbb{S}_{\mathbb{v}\mathbb{u}} = \boldsymbol{Self - regulated\ ConvNet} \oplus \boldsymbol{Lightweighted\ depth\ attention}(\mathbb{V}\mathbb{S}_\mathbb{u}) \quad \langle 10 \rangle$$

Similarly, to obtain context-level features from the set of $\mathbb{N}_{\mathbb{u}\mathbb{C}}$ contextual utterances, the mean value is computed for all frames denoted as $\mathbb{V}_{\mathbb{u}\mathbb{C}}$. In cases where speaker information is present, the utterance $\mathbb{V}_{\mathbb{u}\mathbb{C}}$ is concatenated with the corresponding speaker information $\mathbb{S}_{\mathbb{v}\mathbb{C}}$ defined by **Eq. ⟨10⟩** and **Eq. ⟨11⟩**.

$$\mathbb{V}\mathbb{S}_\mathbb{C} = \big((\mathbb{V}_{\mathbb{u}\mathbb{C}1} \oplus \mathbb{S}_{\mathbb{v}\mathbb{C}1}), (\mathbb{V}_{\mathbb{u}\mathbb{C}2} \oplus \mathbb{S}_{\mathbb{v}\mathbb{C}2}), \ldots\ldots, (\mathbb{V}_{\mathbb{u}\mathbb{C}\mathbb{N}} \oplus \mathbb{S}_{\mathbb{v}\mathbb{C}\mathbb{N}})\big) \quad \langle 11 \rangle$$

$$\mathbb{V}_{\mathbb{u}\mathbb{C}} \oplus \mathbb{S}_{\mathbb{v}\mathbb{C}} = \boldsymbol{Self - regulated\ ConvNet} \oplus \boldsymbol{Lightweighted\ depth\ attention}(\mathbb{V}\mathbb{S}) \quad \langle 12 \rangle$$

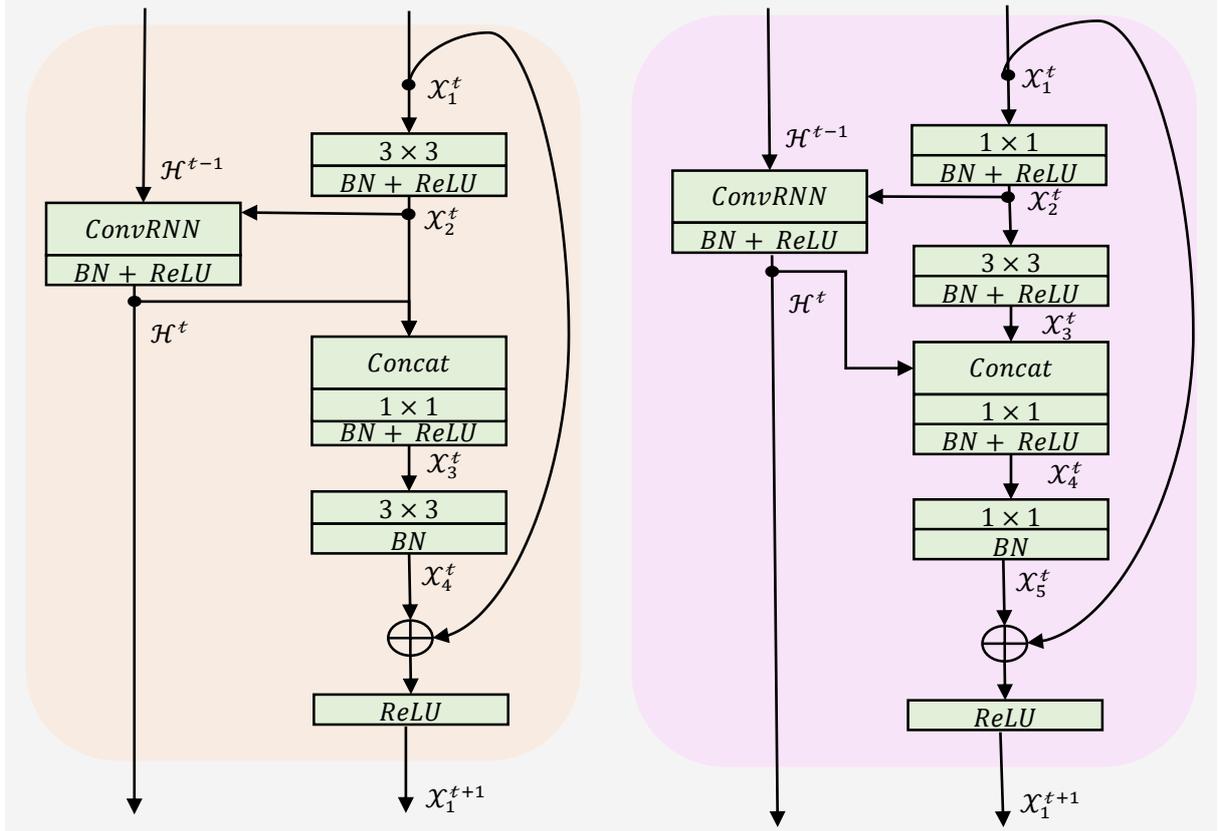

**Figure 4** Self-regulated ConvNet in which $\mathcal{H}$ refers to the hidden states, $\mathcal{X}$ refers to the input feature map, and $t$ denotes the number of building blocks

### 3.3.4 Utterance-Level Feature Extraction from Acoustic Content

Suppose there is a set of $\mathbb{N}_a$ acoustic frames at utterance level denoted as $\mathbb{A}_{1:\mathbb{N}_a} = \{\mathbb{A}_1, \mathbb{A}_2, \ldots\ldots, \mathbb{A}_{\mathbb{N}_a}\}$. Librosa library has been utilized to extract acoustic information. Similar to the process of extracting visual features, the methodology utilised here also involves the computation of the average value of all frames to extract information pertaining to utterances denoted as $\mathbb{A}_u$. In cases where speaker information is present, the utterance $\mathbb{A}_u$ is concatenated with the corresponding speaker information $\mathbb{S}_{au}$ given by **Eq. ⟨13⟩**. The notation $\mathbb{A}_u \oplus \mathbb{S}_{au}$ used is where $\mathbb{A}_u$ belongs to the set of real numbers $\mathfrak{R}$ and has a cardinality of 283. Also, it is important to keep in mind that audio recordings often consist of a variety of speakers, ambient noise, cues for laughter, and other sounds in the background. As a result, the consideration of contextual factors is not integrated into acoustic analysis, as it might cause challenges in distinguishing it from the laughter portion of the conversation. Therefore, an empty list [ ] is employed within the context of acoustic content.

$$\mathbb{A}_u \oplus \mathbb{S}_{au} = \textbf{Librosa tool}\big(\boldsymbol{Concatenation}(\mathbb{A}_u, \mathbb{S}_{au})\big) \qquad \langle 13 \rangle$$

### 3.3.5 Multi-Headed Attention-Based Feature Fusion

In the very first step, all the utterance and context-level features of all the modalities are concatenated together, as represented in **Eq. ⟨14⟩**.

$$\mathbb{G}_{cat} = \boldsymbol{Concatenation}\big((\mathbb{G}_u \oplus \mathbb{S}_{gu}), (\mathbb{G}_c \oplus \mathbb{S}_{gc})\big)$$

$$\mathbb{V}_{cat} = \boldsymbol{Concatenation}\big((\mathbb{V}_u \oplus \mathbb{S}_{vu}), (\mathbb{V}_{uc} \oplus \mathbb{S}_{vc})\big)$$

$$\mathbb{A}_{cat} = \boldsymbol{Concatenation}\big((\mathbb{A}_u \oplus \mathbb{S}_{au}), ([\,])\big) \qquad \langle 14 \rangle$$

Subsequently, $\mathbb{G}_{cat}$, $\mathbb{V}_{cat}$, and $\mathbb{A}_{cat}$ are individually fed into the linear layer followed by the multi-headed attention layer as defined by **Eq. $\langle 15 \rangle$**

$$\mathbb{L}_{\mathbb{G}} = \boldsymbol{Multi-headed\ attention}\big(\boldsymbol{Linear}(\mathbb{G}_{cat})\big)$$

$$\mathbb{L}_{\mathbb{V}} = \boldsymbol{Multi-headed\ attention}\big(\boldsymbol{Linear}(\mathbb{V}_{cat})\big)$$

$$\mathbb{L}_{\mathbb{A}} = \boldsymbol{Multi-headed\ attention}\big(\boldsymbol{Linear}(\mathbb{A}_{cat})\big) \qquad \langle 15 \rangle$$

The features derived from various modalities are concatenated and subsequently fed into a linear layer, which is succeeded by a multi-headed attention layer. This process yields a highly significant multi-modal feature vector, as outlined in **Eq. $\langle 16 \rangle$**.

$$\mathbb{GVA} = \boldsymbol{Multi-headed\ attention}\Big(\boldsymbol{Linear}\big(\boldsymbol{Concatenation}(\mathbb{L}_{\mathbb{G}}, \mathbb{L}_{\mathbb{V}}, \mathbb{L}_{\mathbb{A}})\big)\Big) \qquad \langle 16 \rangle$$

Finally, the softmax layer is utilized to forecast the classification label as either sarcastic or non-sarcastic, as illustrated in **Eq. $\langle 17 \rangle$**.

$$\mathbb{C} = \boldsymbol{Softmax}(\mathbb{GVA}) \qquad \langle 17 \rangle$$

## 4 Experiment and Result

The following subsection covers comprehensive details related to the dataset used throughout the study, the experimental configurations of the proposed methodology, and evaluations of its performance.

### 4.1 Dataset Used

The MUStARD dataset, as provided by [32], is utilized for the purpose of multimodal sarcasm recognition. This dataset comprises a total of 690 utterances, with 345 being sarcastic and 345 being non-sarcastic. The data was gathered from various well-known television series, including Sarcasmaholics Anonymous, Friends, and The Golden Girls. Similar to previous research, we analyse our proposed framework in two distinct experimental setups.

One of the scenarios is the "speaker independent" configuration. The present study employs utterances obtained from the Friends Series as testing data, while the remaining utterances are utilized as training data. The alternative setup is the "speaker dependent", wherein the dataset is partitioned into five folds. During each of the five iterations, the ith fold is designated as the testing set, while the rest of the sample folds are utilized for training purposes. Subsequently, five datasets can be acquired.

In addition to the MUStARD dataset, we have also used the extended version MUStARD++ [36] to execute a cross-dataset study as part of a generalization research to test the resilience of VyAnG-Net. Our proposed approach, VyAnG-Net, was trained using the MUStARD dataset for this experimental investigation, and its performance was evaluated using an unseen MUStARD++ dataset. Consistent with prior research, we utilize $\mathbb{Accuracy}(A)$, $\mathbb{Precision}(P)$, $\mathbb{Recall}(R)$, and $\mathbb{F1\ Score}(F1)$ as the metrics for evaluation. Regarding the "speaker dependent" configuration, the outcomes are presented by computing the mean of the results obtained from five distinct evaluation sets.

### 4.2 Experimental Setup

The proposed method was executed by utilising the Keras and PyTorch framework. The evaluation metrics employed for recognising sarcasm are $\mathbb{Accuracy}(A)$, $\mathbb{Precision}(P)$,

$\mathbb{Recall}(R)$, an $\mathbb{F1\ Score}(F1)$. With respect to all experimental procedures, the architectures employed in this study incorporate, Rectified Linear Unit (ReLU) as an activation function, a dropout rate of 0.4, the Adam optimisation algorithm with a learning rate of 0.001, and a batch size of 32. The training process of the model was conducted for a total of 200 epochs, because the results were not satisfactory beyond this limit i.e., we pertain to a limit of 200 number of epochs. The Adam optimizer is employed in conjunction with Softmax as a classifier to identify sarcasm. Furthermore, the implementation of the sigmoid activation function and the optimisation of binary cross-entropy as the loss function were utilized. The model we have proposed is evaluated using the MUStARD [32] dataset. The results of the grid search were used to obtain the optimal hyper-parameters. The aim of our study is to utilize a consistent hyper-parameter setup across all experimental trials.

Also, it is worth noting that all the experimental procedures were conducted using high-end GPU systems featuring the following specifications: NVIDIA Titan RTX (48 GB), 256 GB of RAM, 10 TB of storage space, and an Intel Xeon Silver 4116 processor. The computational model being analysed demonstrates a relatively low requirement for GPU memory, utilizing approximately 2 gigabytes. Typically, on an average, each epoch requires a duration of approximately 3 to 4 seconds.

In order to ensure a rigorous evaluation in accordance with current cutting-edge frameworks, we conducted comprehensive experiments encompassing unimodal, bimodal, and trimodal approaches for both "speaker dependent" and "speaker independent" setups.

### 4.3 Results and Discussion

The proposed architecture was assessed through a comprehensive analysis of all potential combinations of input. These include unimodal inputs such as $\mathbb{G}$, $\mathbb{V}$, and $\mathbb{A}$, bimodal inputs such as $\mathbb{G} \oplus \mathbb{V}$, $\mathbb{V} \oplus \mathbb{A}$, and $\mathbb{G} \oplus \mathbb{A}$, as well as trimodal input $\mathbb{G} \oplus \mathbb{V} \oplus \mathbb{A}$. In the context of speaker dependent configuration, our proposed VyAnG-Net (for trimodal) proved its superior performance (**Table 2**) with a $\mathbb{Precision}(P)$ of 78.83% (an increase of 6.93 [32], 3.63[35], and 4.63[36] points), $\mathbb{Recall}(R)$ of 78.21% (an increase of 6.81 [32], 3.61[35], and 4.01[36] points), and $\mathbb{F1} - \mathbb{Score}(F1)$ of 78.52% (an increase of 7.02 [32], 4.02[35], and 4.32[36] points). Experimental evidence indicates that the trimodal approach outperforms both the unimodal and bimodal approaches.

In addition, for speaker independent configuration, the proposed model VyAnG-Net (for trimodal) exhibited exceptional performance (**Table 3**) with $\mathbb{Precision}(P)$ of 75.69% (an increase of 11.39 [32], 4.39[35], and 3.59[36] points), $\mathbb{Recall}(R)$ of 75.52% (an increase of 12.92 [32], 4.22[35], and 3.52[36] points), and $\mathbb{F1\ Score}(F1)$ of 75.6% (an increase of 12.8[32], 5.6[35], and 3.6[36] points).

**Figure 5** and **Figure 6** illustrates the curves pertaining to testing loss, accuracy, precision, recall, and F1 scores for speaker dependent and speaker independent configuration.

Table 2 Experimental results for *speaker dependent* setup using VyAnG-Net

| Modalities | Speaker dependent | | | |
|---|---|---|---|---|
| | $\mathbb{Accuracy}(A)$ | $\mathbb{Precision}(P)$ | $\mathbb{Recall}(R)$ | $\mathbb{F1\ Score}(F1)$ |
| $\mathbb{G}$ (**unimodal**) | 73.16 | 72.53 | 72.4 | 72.45 |
| $\mathbb{V}$ (**unimodal**) | 73.81 | 72.93 | 71.86 | 72.4 |
| $\mathbb{A}$ (**unimodal**) | 74.42 | 72.7 | 72.69 | 72.7 |
| $\mathbb{G} \oplus \mathbb{V}$ (**bimodal**) | 75.68 | 74.94 | 74.26 | 74.61 |

| Modalities | Speaker dependent | | | |
|---|---|---|---|---|
| | Accuracy(**A**) | Precision(**P**) | Recall(**R**) | F1 Score(**F1**) |
| $\mathbb{V} \oplus \mathbb{A}$ (**bimodal**) | 77.37 | 77.53 | 77.45 | 77.49 |
| $\mathbb{G} \oplus \mathbb{A}$ (**bimodal**) | 77.14 | 76.84 | 76.92 | 76.87 |
| $\mathbb{G} \oplus \mathbb{V} \oplus \mathbb{A}$ (**trimodal**) | **79.86** | **78.83** | **78.21** | **78.52** |

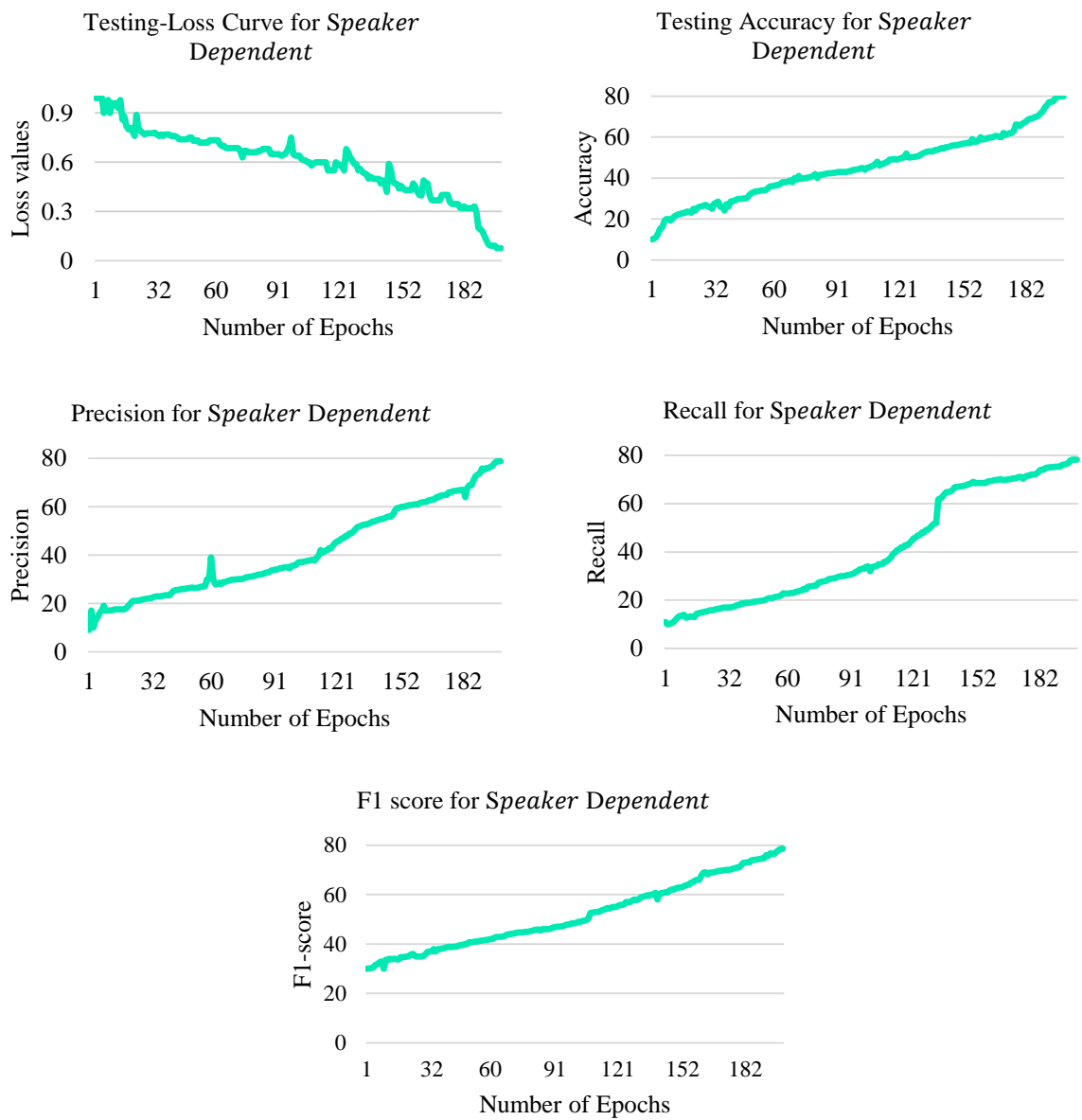

**Figure 5** Testing curve for loss, accuracy, precision, recall, and F1 scores for *speaker dependent*

**Table 3** Experimental results for speaker independent setup using VyAnG-Net

| Modalities | Speaker Independent | | | |
|---|---|---|---|---|
| | Accuracy(**A**) | Precision(**P**) | Recall(**R**) | F1 Score(**F1**) |
| $\mathbb{G}$ (**unimodal**) | 69.47 | 68.36 | 68.24 | 68.29 |
| $\mathbb{V}$ (**unimodal**) | 70.15 | 70.89 | 70.16 | 70.52 |
| $\mathbb{A}$ (**unimodal**) | 70.92 | 71.12 | 71.23 | 71.17 |
| $\mathbb{G} \oplus \mathbb{V}$ (**bimodal**) | 72.42 | 72.64 | 72.61 | 72.61 |

| Modalities | Speaker Independent | | | |
|---|---|---|---|---|
| | Accuracy(**A**) | Precision(**P**) | Recall(**R**) | F1 Score(**F1**) |
| $\mathbb{V} \oplus \mathbb{A}$ (**bimodal**) | 74.74 | 74.51 | 74.32 | 74.41 |
| $\mathbb{G} \oplus \mathbb{A}$ (**bimodal**) | 74.64 | 73.96 | 73.48 | 73.72 |
| $\mathbb{G} \oplus \mathbb{V} \oplus \mathbb{A}$ (**trimodal**) | **76.94** | **75.69** | **75.52** | **75.6** |

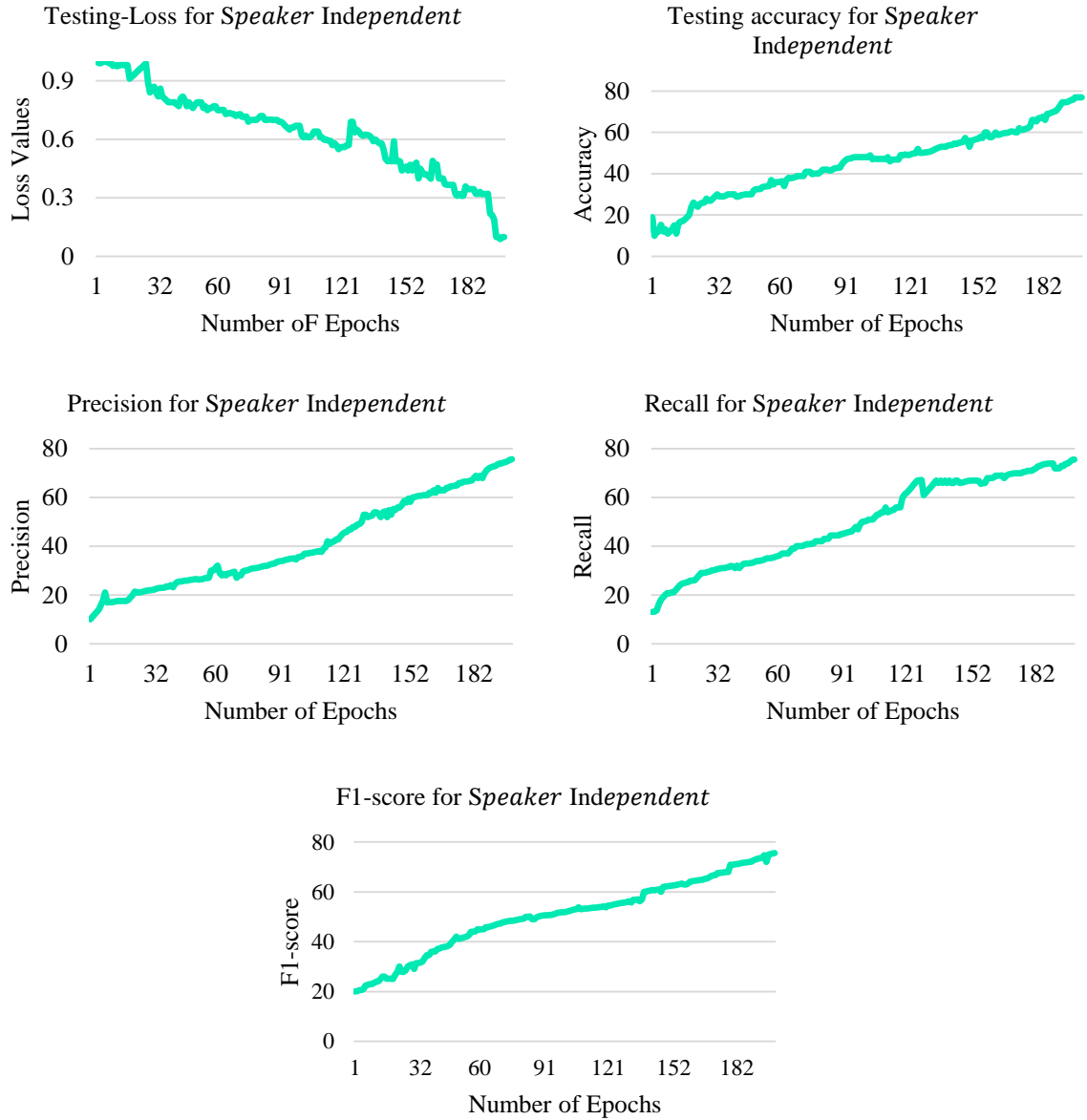

**Figure 6** Testing curve for loss, accuracy, precision, recall, and F1 scores for *speaker independent*

### 4.4 Comparative Analysis with the Baselines

In this study, we conducted a comparative analysis using consistent experimental conditions with the pre-existing models, namely Baseline − 1 [32], Baseline − 2 [35], Baseline − 3 [36], Baseline − 4 [37], and Baseline − 5 [33], which were developed using a similar MUStARD [32] dataset.

Baseline − 1 [32] was the first to release and work on the video dataset MUStARD in the field of MSR. This study employed ResNet-152 for the purpose of extracting visual features, BERT-based uncased architecture for extracting textual features, and the librosa library for extracting auditory features. Finally, a Support Vector Machine (SVM) was employed as a classifier to distinguish between instances of sarcasm and non-sarcasm. Later, the IWAN framework was proposed in Baseline − 2 [35], which incorporates attention mechanisms and employs similar architectures as employed in [32] for extracting visual and textual features. However, for the extraction of auditory features, the OpenSmile tool was utilized. The authors in Baseline − 3 [36] proposed a novel methodology for MSR that integrates ResNet-152, BART, and librosa for the purpose of feature extraction across various modalities. This research study has undertaken the task of emotion recognition in addition to implicit sentiment, specifically sarcasm. The study conducted by Baseline − 5 [33] employed comparable architectures as those utilised in the previous research [32], in conjunction with a late fusion strategy, to detect instances of sarcasm in utterances. In addition to the aforementioned research studies, it is noteworthy to mention that Baseline − 4 [37] was the pioneer in utilising fuzzy logic and quantum theory for the purpose of detecting sarcasm in video utterances.

Our methodology outperforms all of the baseline models that have been addressed in **Table 4** and **Table 5**. The results indicate that the VyAnG-Net model is advantageous for MSR due to its ability to comprehensively stimulate the relationship between all the modalities at a more profound level. This is achieved through the use of a glossary branch that employs an attention-based tokenization approach and a dedicated attention module to identify the most salient features from the video frames, along with a multi-headed attention-based feature fusion technique. The visualisation of VyAnG-Net on MUStARD dataset versus the cutting-edge baseline approaches in terms of Accuracy, Precision, Recall, and F1 scores for speaker dependent and speaker independent configuration are presented in **Figure 7** and **Figure 8**.

**Table 4** Comparison with baseline models for speaker dependent setup

| Modalities | Precision(P) | | | | | | Recall(R) | | | | | | F1 Score(F1) | | | | | |
|---|---|---|---|---|---|---|---|---|---|---|---|---|---|---|---|---|---|---|
| | VyAnG-Net | [32] | [35] | [36] | [37] | [33] | VyAnG-Net | [32] | [35] | [36] | [37] | [33] | VyAnG-Net | [32] | [35] | [36] | [37] | [33] |
| $\mathbb{G}$ | **72.53** | 65.1 | - | - | - | 67.73 | **72.4** | 64.6 | - | - | - | 66.8 | **72.45** | 64.6 | - | - | - | 67.66 |
| $\mathbb{V}$ | **72.93** | 68.1 | - | - | - | 71.6 | **71.86** | 67.4 | - | - | - | 71.06 | **72.4** | 67.4 | - | - | - | 70.84 |
| $\mathbb{A}$ | **74.71** | 65.9 | - | - | - | 73.22 | **73.69** | 64.6 | - | - | - | 72.61 | **73.7** | 64.6 | - | - | - | 72.44 |
| $\mathbb{G} \oplus \mathbb{V}$ | **74.94** | 72 | 69.1 | - | - | 73.44 | **74.26** | 71.6 | 68.9 | - | - | 73.33 | **74.61** | 71.6 | 68.9 | - | - | 73.3 |

| Modalities | Precision(P) | | | | | | Recall(R) | | | | | | F1 Score(F1) | | | | | |
|---|---|---|---|---|---|---|---|---|---|---|---|---|---|---|---|---|---|---|
| | VyAnG-Net | [32] | [35] | [36] | [37] | [33] | VyAnG-Net | [32] | [35] | [36] | [37] | [33] | VyAnG-Net | [32] | [35] | [36] | [37] | [33] |
| V⊕A | **77.53** | 66.2 | - | - | - | 73.81 | **77.45** | 65.7 | - | - | - | 73.33 | **77.49** | 65.7 | - | - | - | 73.19 |
| G⊕A | **76.84** | 66.6 | 70.8 | - | - | 71.19 | **76.92** | 66.2 | 70.2 | - | - | 70.87 | **76.87** | 66.2 | 70.2 | - | - | 70.87 |
| G⊕V⊕A | **78.83** | 71.9 | 75.2 | 74.2 | 75.3 | 73.8 | **78.21** | 71.4 | 74.6 | 74.2 | 75.5 | 73.62 | **78.52** | 71.5 | 74.5 | 74.2 | 75.4 | 73.58 |

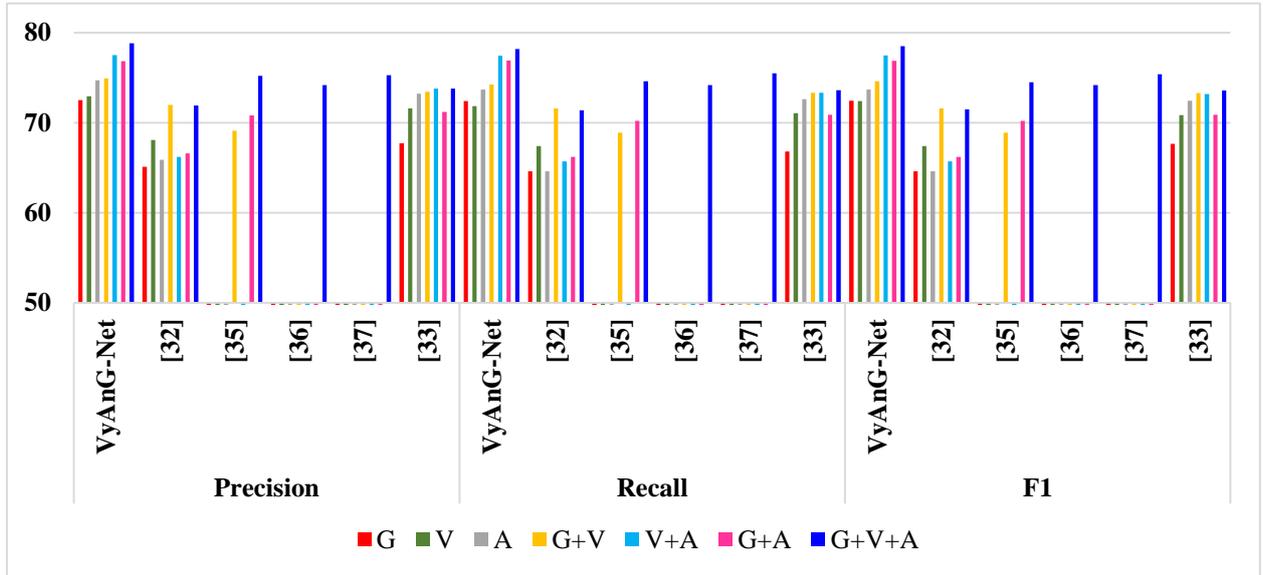

**Figure 7** Evaluation of our proposed framework VyAnG-Net on MUStARD dataset versus the cutting-edge baseline approaches in terms of *Accuracy*, *Precision*, *Recall*, and *F*1 *scores* for *speaker dependent* configuration

**Table 5** Comparison with baseline models for *speaker independent* setup

| Modalities | Precision(**P**) | | | | | Recall(**R**) | | | | | F1 Score(**F1**) | | | | |
|---|---|---|---|---|---|---|---|---|---|---|---|---|---|---|---|
| | VyAnG-Net | [32] | [35] | [36] | [37] | [33] | VyAnG-Net | [32] | [35] | [36] | [37] | [33] | VyAnG-Net | [32] | [35] | [36] | [37] | [33] |
| G | **68.36** | 60.9 | - | - | - | 58.18 | **68.24** | 59.6 | - | - | - | 57.75 | **68.29** | 59.8 | - | - | - | 57.84 |
| V | **70.89** | 54.9 | - | - | - | 70.37 | **70.16** | 53.4 | - | - | - | 70.56 | **70.52** | 53.6 | - | - | - | 70.13 |

| Modalities | Precision (P) | | | | | | Recall (R) | | | | | | F1 Score (F1) | | | | | |
|---|---|---|---|---|---|---|---|---|---|---|---|---|---|---|---|---|---|---|
| | VyAnG-Net | [32] | [35] | [36] | [37] | [33] | VyAnG-Net | [32] | [35] | [36] | [37] | [33] | VyAnG-Net | [32] | [35] | [36] | [37] | [33] |
| $\mathbb{A}$ | 73.92 | 65.1 | - | - | - | 72.93 | 73.23 | 62.6 | - | - | - | 71.4 | 73.58 | 62.7 | - | - | - | 71.21 |
| $\mathbb{G}\oplus\mathbb{V}$ | 72.64 | 62.2 | 63.2 | - | - | 64.7 | 72.61 | 61.5 | 63.1 | - | - | 64.72 | 72.61 | 61.5 | 63.1 | - | - | 63.98 |
| $\mathbb{V}\oplus\mathbb{A}$ | 74.51 | 64.1 | - | - | - | 72.44 | 74.32 | 61.8 | - | - | - | 71.57 | 74.41 | 61.9 | - | - | - | 70.27 |
| $\mathbb{G}\oplus\mathbb{A}$ | 73.96 | 64.7 | 59.6 | - | - | 62.75 | 73.48 | 62.9 | 60.0 | - | - | 61.24 | 73.72 | 63.1 | 59.7 | - | - | 61.33 |
| $\mathbb{G}\oplus\mathbb{V}\oplus\mathbb{A}$ | 75.69 | 64.3 | 71.9 | 72.1 | 75.3 | 71.55 | 75.52 | 62.6 | 71.3 | 72 | 75.5 | 71.52 | 75.6 | 62.8 | 70.0 | 72 | 75.4 | 70.99 |

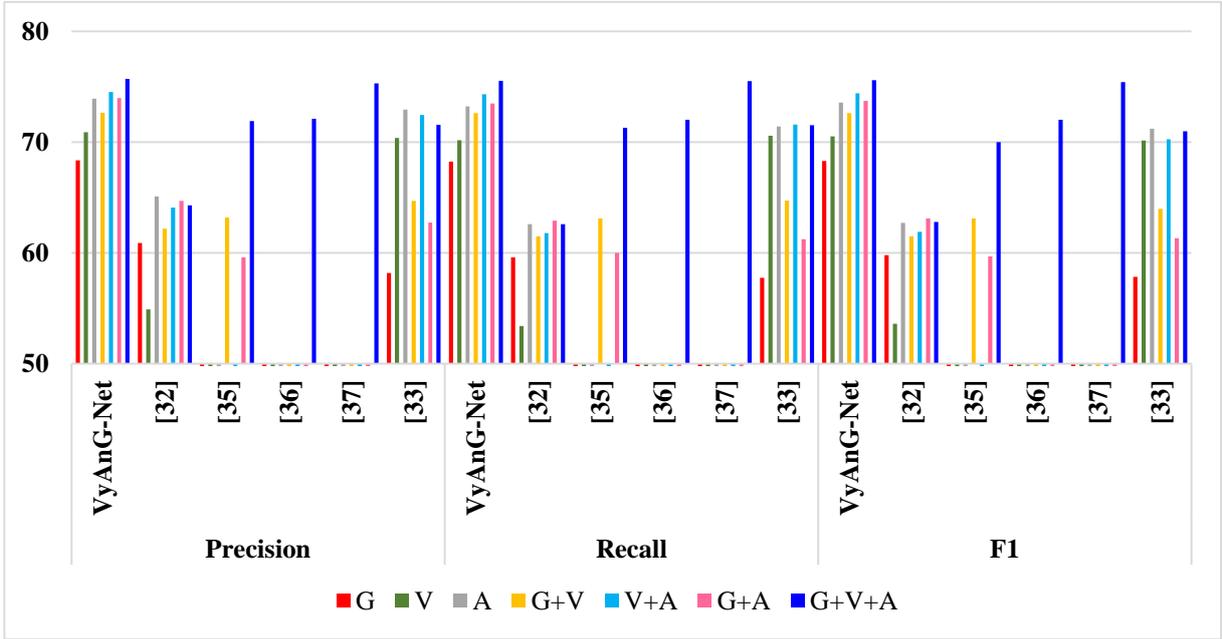

**Figure 8** Evaluation of our proposed framework VyAnG-Net on MUStARD dataset versus the cutting-edge baseline approaches in terms of *Accuracy*, *Precision*, *Recall*, and *F1 scores* for *speaker independent* configuration

In this section, a series of ablation trials were carried out on the MUStARD dataset to more accurately evaluate the effectiveness of each proposed module specifically for trimodal ($\mathbb{G}\oplus\mathbb{V}\oplus\mathbb{A}$). The VyAnG-Net framework was employed to generate three distinct variants, namely "VyAnG-Net w/o attention-seeking tokenizer", "VyAnG-Net w/o lightweight depth attention", and "VyAnG-Net w/o multi-headed attention" as illustrated in **Table 6** and

**Table 7**. The aforementioned modifications entail the extraction of features from the textual content of video utterance's subtitles by eliminating the attention-seeking tokenizer in the glossary branch module, removal of the lightweight depth attention module in the visual branch, and instead of utilising multi-headed attention for fusion, a method of directly concatenating the obtained feature representation from multiple branches was employed. **Table 6** and **Table 7** summarise the outcomes obtained from the ablation experiments.

Based on these observations, the following conclusions can be drawn: The VyAnG-Net, as the proposed model, encompasses all modules and exhibits superior performance on the MUStARD dataset. The removal of a single module would lead to inadequate predictive outcomes. Based on these observations, it can be inferred that every proposed module is crucial and plays a significant role in the overall performance. Also it can be observed that in the speaker dependent configuration, the proposed model is performing better because of the intermixing of variety of videos in the training set as compared to the speaker independent configuration, resulting in enhanced generalizability of speaker dependent configuration.

**Table 6** The results of the ablation trials carried out on the MUStARD dataset for a *speaker dependent* configuration

| Methods | MUStARD | | | |
|---|---|---|---|---|
| | Accuracy($A$) | Precision($P$) | Recall($R$) | F1 Score($F1$) |
| VyAnG-Net w/o attention-seeking tokenizer | 76.49 | 75.87 | 75.62 | 75.74 |
| VyAnG-Net w/o lightweight depth attention | 75.24 | 75.53 | 74.97 | 75.47 |
| VyAnG-Net w/o multi-headed attention | 73.86 | 72.63 | 72.2 | 72.41 |
| **VyAnG-Net** | **79.86** | **78.83** | **78.21** | **78.52** |

**Table 7** The results of the ablation trials carried out on the MUStARD dataset for a *speaker independent* configuration

| Methods | MUStARD | | | |
|---|---|---|---|---|
| | Accuracy($A$) | Precision($P$) | Recall($R$) | F1 Score($F1$) |
| VyAnG-Net w/o attention-seeking tokenizer | 73.96 | 73.32 | 72.84 | 73.08 |
| VyAnG-Net w/o lightweight depth attention | 71.42 | 72.03 | 71.99 | 72.00 |
| VyAnG-Net w/o multi-headed attention | 73.85 | 72.74 | 72.62 | 72.68 |
| **VyAnG-Net** | **76.94** | **75.69** | **75.52** | **75.6** |

## 6 Generalization study

In the recent years the MUStARD dataset has shown remarkable results for various emerging multimodal sarcasm recognition frameworks; however, these approaches do not possess the generalizability that algorithms need to examine samples from other domains or datasets. They are more concerned in conducting thorough architectural in-dataset analyses. Consequently, rather than limiting ourselves to assessing VyAnG-Net on a single dataset, we provide a cross-dataset evaluation that puts a model trained on one dataset to the test on another.

Therefore, in addition to the experiment discussed in the above sections, we undertake a cross-dataset study as part of a generalization research to test the resilience of VyAnG-Net. Our proposed approach, VyAnG-Net, was trained using the MUStARD dataset for this experimental investigation, and its performance was evaluated using the MUStARD++ dataset.

Throughout the training stage of the cross-dataset study, 80% of the samples from the MUStARD dataset were chosen at random for training, while 10% were allocated for validation. For the testing phase, 10% of the samples had been picked at random from the MUStARD++ dataset to assess the predictive power of the proposed method using various parameters including $\mathbb{Accuracy}(A)$, $\mathbb{Precision}(P)$, $\mathbb{Recall}(R)$, $\mathbb{F1\ Score}(F1)$.

The efficiency of VyAnG-Net, which is based on lightweight depth attention module and the attention-based tokenization approach, was tested through cross-dataset study. From **Table 8** It is evident that our technique could identify instances from datasets other than the ones it was trained on. Hence, it can be proved that our suggested model, VyAnG-Net, is effective, generalizable and more reliable than earlier state-of-the-art solutions.

Table 8 Evaluating the resilience of VyAnG-Net using cross-dataset study that uses MUSTaRD dataset for training and validation, whereas MUStARD++ is employed for testing purposes.

| Dataset utilized | VyAnG-Net | | | |
|---|---|---|---|---|
| | $\mathbb{Accuracy}(A)$ | $\mathbb{Precision}(P)$ | $\mathbb{Recall}(R)$ | $\mathbb{F1\ Score}(F1)$ |
| MUStARD (for training and validation) and MUStARD ++ (for testing) | 73.93 | 72.41 | 72.05 | 72.23 |

## 7 Conclusion & Future Scope

This paper proposes a novel VyAnG-Net for multi-modal sarcasm recognition by uncovering visual, acoustic and glossary features. Our proposed methodology analyses the interaction between all three modalities more effectively than earlier innovative methods. The sentiment evoked by a particular phrase may vary across different contexts. Therefore, it is vital to utilise auditory and visual cues to boost prediction accuracy. Motivated by this, we have introduced an innovative visual branch incorporating a lightweight depth attention module to extract the most salient features from video frames.

Additionally, a glossary branch utilises an attention-based tokenization approach to capture the most critical contextual features from the textual content provided by video subtitles. Furthermore, an utterance-level feature extraction method for acoustic content has been implemented along with a multi-headed attention-based feature fusion technique to combine features obtained from each of the distinct modalities. Our extensive experiments on the publicly available dataset MUStARD indicate that our proposed model outperforms other competitive baseline models.

Despite attaining promising results, there are still numerous avenues open for further research.

**Fusion Strategy:** Our research has focused on the feature fusion approach by utilising a multi-headed attention mechanism for MSR. Subsequent research endeavours may explore advanced spatiotemporal fusion techniques to effectively capture the correlation between different modalities, such as tensor-based fusion. A potential alternative strategy entails formulating more fusion methodologies that can better encapsulate the discrepancies among diverse modalities to identify occurrences of sarcasm with greater efficacy.

**Neural Baseline:** Further research studies should aim to implement transfer learning, pre-training, low-parameter, or domain adaptation models as potential solutions. Also, the architectures with fewer trainable parameters are more advantageous for facilitating real-time deployment.

**Visual sarcasm recognition:** So far, there has been a lack of research in the domain of sarcasm detection utilising solely visual cues that include the embedded text, primarily due to the absence of an appropriate dataset. Therefore, developing a visual sample corpus centred on memes is essential to work effectively in this field.

**Integration of Attention-based strategies:** In the modern era, there has been an explosion of attention modules that need further study for their potential integration with diverse forms of Convolutional Neural Networks (ConvNets) and Recurrent Neural Networks (RNNs). This integration could facilitate the efficient extraction of pertinent information from both textual and visual inputs.